\documentclass[10pt,twocolumn,letterpaper]{article}

\usepackage{iccv}
\usepackage{times}
\usepackage{epsfig}
\usepackage{graphicx}
\usepackage{amsmath}
\usepackage{amssymb}

\usepackage{multirow}

\usepackage{latexsym}
\usepackage{amssymb,amsmath,amsfonts}
\usepackage{algorithmic}
\usepackage{nopageno}

\iccvfinalcopy 

\def\iccvPaperID{****} 

\usepackage[ruled]{algorithm2e}

\newcounter{defcounter}
\newenvironment{myequation}{%
	\addtocounter{equation}{-1}
	\refstepcounter{defcounter}
	
	\begin{equation}}
	{\end{equation}}

\newtheorem{mydef}{Definition}

\begin{document}

\title{Manifold Constrained Low-Rank Decomposition}

\author{Chen Chen$^2$, Baochang Zhang$^{1,3,*}$, Alessio Del Bue$^4$, Vittorio Murino$^4$\\
$^1$State Key Laboratory of Satellite Navigation System and Equipment Technology, Shijiazhuang, China\\
$^2$Center for Research in Computer Vision (CRCV), University of Central Florida (UCF)\\
$^3$School of Automation Science and Electrical Engineering, Beihang University, Beijing, China\\
$^4$ Istituto Italiano di Tecnologia, Genova, Italy\\
{\tt\small chenchen870713@gmail.com, alessio.delbue@iit.it, bczhang@buaa.edu.cn, vittorio.murino@iit.it}
\thanks{Baochang Zhang is the corresponding author.}
}

\maketitle

\begin{abstract}
Low-rank decomposition (LRD) is a state-of-the-art method for visual data reconstruction and modelling. However, it is a very challenging problem when the image data contains significant occlusion, noise, illumination variation, and misalignment from rotation or viewpoint changes. We leverage the specific structure of data in order to improve the performance of LRD when the data are not ideal. To this end, we propose a new framework that embeds manifold priors into LRD. To implement the framework, we design an alternating direction method of multipliers (ADMM) method which efficiently integrates the manifold constraints during the optimization process. The proposed approach is successfully used to calculate low-rank models from face images, hand-written digits and planar surface images. The results show a consistent increase of performance when compared to the state-of-the-art over a wide range of realistic image misalignments and corruptions.
\end{abstract}

\section{Introduction}
	
	With the increasing number of images and videos produced everyday, it becomes more problematic when existing algorithms have to deal with realistic data containing severe occlusion, misalignment, noise, significant illumination variation, and viewpoint changes \cite{Chenglong,liang2016supervised,JiangSR}.
	Low-rank decomposition (LRD) techniques have been an important tool in batch data analysis in the past decade, which effectively converts high dimensional raw data into a compact and low-dimensional representation. It has been successfully used in a variety of applications such as subspace segmentation \cite{47}, visual tracking, image clustering \cite{49} and video background foreground separation \cite{1}. However, this technique works properly when the data is captured in an ideal situation or it is manually aligned. The performance of the algorithm degrades significantly in case of rotation, corruption, occlusion and misalignment in the data. In such situations, low-rank matrices cannot be accurately recovered from the data because geometrical distortions are difficult to grasp with a linear subspace model.
	
	To make LRD based methods applicable in more
	realistic scenarios, various solutions have been developed. For instance, a sophisticated measure of image similarity is used in \cite{linzhouchen,similarity1} to address the batch image alignment problem. Alternatively, Learned-Miller's congealing algorithm \cite{similarity1} seeks an alignment that minimizes the sum of entropy of pixel value at each pixel location in the batch of aligned images. Instead of using the entropy, the least squares congealing procedure \cite{similarity2} minimizes the sum of squared distances between pairs of images, and therefore requires the columns to be nearly constant.
	In \cite{similarity3}, Vedaldi \textit{et al.} choose to minimize a log-determinant measure that can be viewed as a smooth surrogate for the rank function.
	The Robust Principal Component Analysis
	(RPCA) algorithm fits a low-rank model, and uses a fitting function to reduce
	the influence of corruptions and occlusions.

	\begin{table*}[t!]
		\caption{A brief description of variables used in the paper.}
		\centering
		\begin{tabular}{|ll|}
			\hline
			 $V_d$: input data matrix (samples in rows)  & $V_r$: low-rank matrix calculated from $V_d$ \\
			 $\tau$: geometric transformation &  $\Delta\tau$: to calculate new $\tau$  \\
			$V_m$: calculated by $\Delta\tau$ and $V_d$ &   $V'_{m}$: embedding of $V_m$  \\
			 $E$: the error matrix  &  $S_\alpha[x]$: soft-thresholding function  \\	
			\hline
		\end{tabular}
	\end{table*}

	Differently, the Robust Alignment by Sparse and Low-rank Decomposition (RASL) \cite{rasl} has shown the potential to solve realistic problems with misalignments and corruptions by using a nuclear-norm minimization based on the alternating direction method of multipliers (ADMM). The core idea of the method is to find an optimal set of transformations such that the matrix of transformed images can be decomposed as a low-rank matrix of recovered aligned images and a sparse matrix of errors. The algorithm is subject to a set of linear equality constraints, which impose a linear relationship with the input data. However, the fact that input data has generally a \textit{nonlinear} structure, \ie distributed over a manifold, is not fully investigated in the optimization process.
	
	In this paper, we provide new insights into the nuclear-norm minimization method, in particular a relevant intuition that was neglected in previous work. That is, data often lies on specific manifolds \cite{ruiping}, especially when the data comes from a well-defined object from a given set of samples (\eg faces, digits, etc.).  From the optimization perspective, assuming that the solution of the optimization problem is always data related, the constraints derived from the data structure can make the algorithm immune to the variations in the testing data \cite{iconip}. Consequently, it is important to incorporate the data structure prior in the learning procedure. 
	In this paper, we show that there is a solution with high practicability that can include manifold constraints in ADMM, which is applied to solve the LRD problem. Technically, we avoid complex nonlinear optimization over the manifold by recasting the problem as a simpler matrix projection over the same manifold. Different from other works \cite{DelBue:etal:PAMI2012}, the manifold is not given but actually estimated from the data, leading to a solution complying with the intrinsic data distribution. 
	
	In summary, the contributions of this paper are twofold. 
	
	$\bullet$ We propose to incorporate the manifold constraints in low-rank decomposition methods, achieving much better results than the prior art.
	
	$\bullet$ We present a manifold embedding based ADMM (MeADMM) framework,  where the manifold constraint is translated into a matrix projection operation computed by a neighbor-preserving embedding process, which greatly simplifies the optimization.
	
	For clarity, we summarize all the variables in Table 1. The matrix $V_d$ is the data matrix containing the data samples at each row. 
	The matrix $V_r$ is the low-rank matrix calculated from $V_d$.
	The geometric transformation functions $\tau$ and $\Delta\tau$ are used to calculate $V_m$ from $V_r$. Using $\tau$ and $\Delta\tau$, we register the data in a way that the rank properties are preserved. Finally, $V'_{m}$  is a manifold embedding of the registered input data $V_m$.
	
	
	\section{Related work}	
	Our work is related to the RASL framework \cite{rasl} and manifold methods. Therefore, the literature overview focuses on RASL methodology as well as relevant manifold approaches.
	\textbf{RASL.} The misalignment problem is one of the most difficult problems in computer vision. By formulating the batch image alignment as searching for a set of transformations that minimize the rank of the transformed images, RASL investigates the linearly correlated relationship among the input images, which is shown in Problem 1 (P1):
	\begin{myequation}
		    \begin{aligned}
		& \hat{V_r},\hat{\tau}=\arg\min & & rank(V_r) + \lambda * ||E||_1, \\
		& \textit{subject to} & & \hspace{-0.25in}  V_d \circ \tau = V_r + E, \hspace{0.1in} \tau \in\mathcal{G}
			 \end{aligned}
	\end{myequation}
	As a practical example, each row of $V_d$ can correspond to an $M \times N$ image frame of a video with $B$ frames while $V_r$ contain a compact low-rank description of the video. In particular, $V_d$ can be a collection of images with variations including rotation, illumination changes, occlusion and geometric transformations given by $\tau \in\mathcal{G}$ where the operator \emph{$\mathcal{G}$} is defined as a $3 \times 3$ matrix \cite{rasl}.
	To efficiently solve the problem, a linearization process is used such that:
	\begin{equation}
		\label{Def_VM1}
		V_d \circ( \tau  +\Delta\tau )= V_d \circ \tau + \sum^B_i J_i\Delta\tau_i \epsilon_i,
	\end{equation}
	where $( \tau  +\Delta\tau )$ gives at each step the new $\tau$ during iterations while the increment $\Delta\tau$ is derived as detailed in  \cite{rasl}.  $J_i$ is the Jacobian matrix of the $i^{th}$ image with respect to the transformation parameters and $\epsilon_i$ denotes the standard bases. The above linearisation process only holds locally. Therefore, linearisation of current estimates is repeated by solving a sequence of convex problems. After the linearisation, a semi-definite programming problem is solved in thousands or millions of variables. Thanks to recent works on high-dimensional nuclear norm minimization, such problems are well within the capability of a standard PC \cite{rasl}.

	
	\textbf{Manifolds.} Manifolds are popular in machine learning, because they allow to describe the intrinsic distribution of data in the Euclidean space.  Most of the existing works related to manifolds focus on modeling the nonlinearity of data. To represent high-dimensional data, manifold learning \cite{isomap} projects the original data onto lower dimensions such that its inherent structure can be preserved. As another application of manifold learning, an embedding of a sample can be obtained by projecting onto a well-designed manifold \cite{Sam:Science}. To exploit the geometry of the marginal distribution, a semi-supervised framework based on manifold regularization is used to learn from both labeled and unlabeled data 
	in the form of a multiple kernel learning. In \cite{ruiping1}, by representing the covariance matrix as a point on a manifold, a new metric is learned for that manifold. Differently, \cite{DelBue:etal:PAMI2012} imposes the manifold constraints in an augmented Lagrange multipliers (ALM) strategy by using a matrix projection as a constraint for the optimized variable, which efficiently computes the solution over several given manifolds. The work leads to a new framework using given manifolds to solve the optimization problem. However, it fails to explain why the variable should stop on a manifold.
	
	Unlike the existing works, we present a new method that exploits learned manifold constraints in the ADMM framework. Instead of empirically adding manifold constraints on a variable, we introduce a manifold based ADMM approach to regulate the optimization problem for LRD. 

	\section{Low-rank decomposition based on manifold constraints}
	A constrained learning model allows to incorporate domain-specific knowledge to balance the learned model based on the implicit structure of the data \cite{ChangRaRo07}. From a machine learning perspective, it is important to simplify the learning stage while improving the accuracy of the solution.
	In this section, we present how manifold constraints can be embedded into an optimization problem. We formulate the LRD optimization problem in terms of MeADMM, resulting in a relaxed and more efficient solution to the new problem defined as P2. 
	
\begin{figure}
		\centering
		\includegraphics[width=0.9\columnwidth]{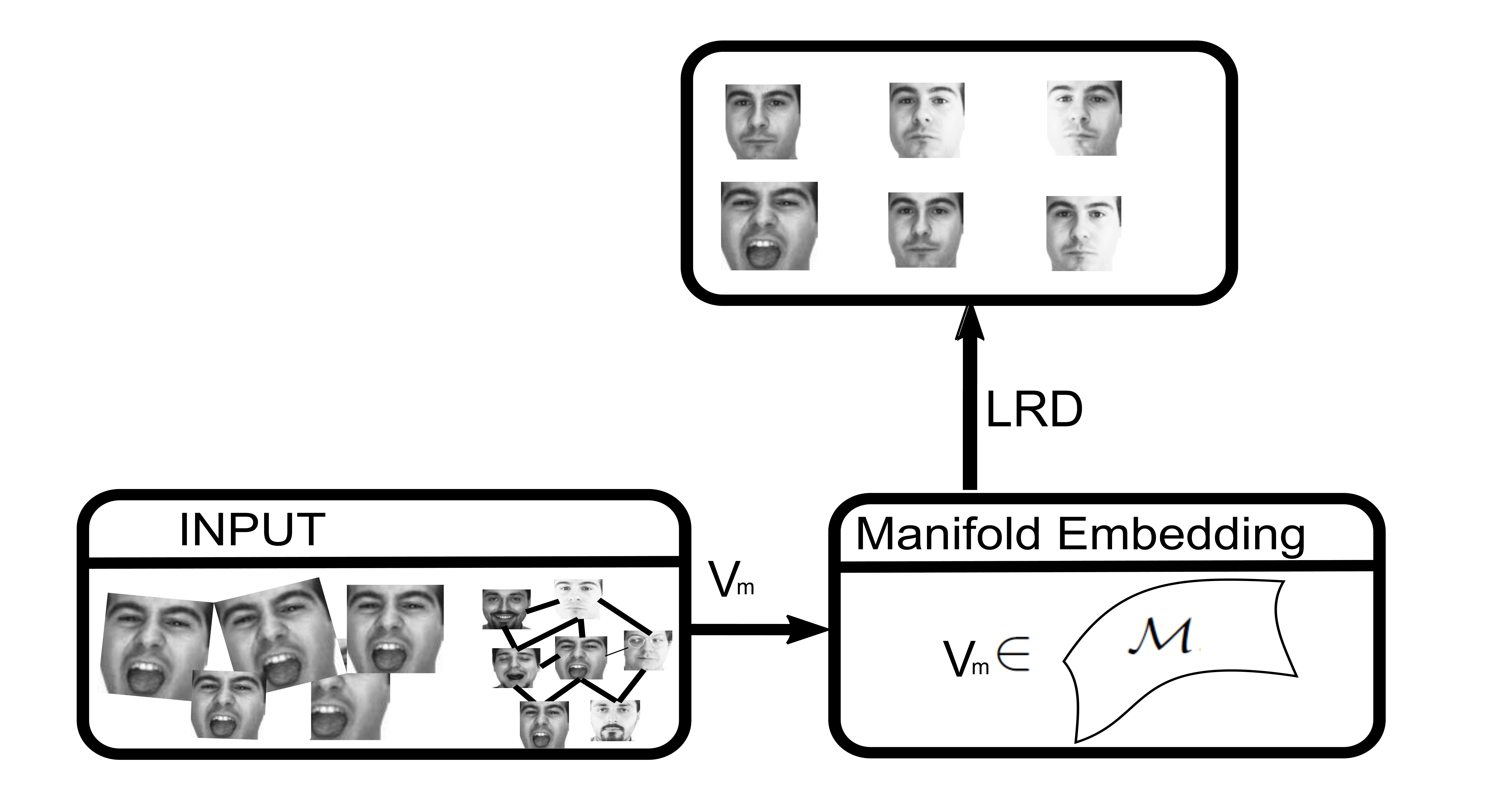}
		
		\caption{The framework of the proposed manifold constrained low-rank decomposition.}
		\label{framework}
	\end{figure}

	Our idea is intuitively illustrated in Fig. \ref{framework}, where the input images are first embedded into a manifold and the low-rank results are obtained afterwards by LRD.
	
	\subsection{LRD reformulation based on MeADMM}	
	
	To efficiently calculate low-rank from data with a nonlinear structure, MeADMM reformulate the manifold constraint in the optimization process. 
	We first introduce a new variable $V_m$ such that $V_m=V_d \circ( \tau  +\Delta\tau )$.
	the matrix $V_m$  replaces $ V_d \circ \tau$, which is a new variable and linearly correlated to $V_r$ in the new problem. LRD is then reformulated as:
	\begin{myequation}
	\small
		\label{p_manifold}
		\begin{aligned}
			& \hat{V_r},\hat{E},\hat{\tau}=\arg\min \{ rank(V_r) + \lambda * ||E||_1\}, \\
			& \textit{subject to} \hspace{0.05in}  V_m = V_r + E, \hspace{0.05in}  V_{r,i}\in\mathcal{M},
			\hspace{0.05in} \tau \in\mathcal{G}
		\end{aligned}
	\end{myequation}
	Normally, if $V_m$ contains images, only a small fraction of pixels will be affected by partial occlusions or corruptions, thus $E$ is considered to be sparse. 
	Supposed that the input data $V_m$ is generally of nonlinear structure, the samples of $V_r$ are reasonably considered to be from a manifold $\mathcal{M}$. That is, $ V_{r,i}\in\mathcal{M}$.  
	We propose to solve the problem in three steps. We first exploit the ALM framework in this subsection to solve the problem without taking manifold constraints into account. In the second step, we introduce the manifold constraint into the objective function in Sec. 3.2. Finally, we solve all variables in Algorithms 1 and 2 in Sec. 3.4.
	
	The idea of ALM is searching for a saddle point of the augmented Lagrangian function instead of directly solving the constrained optimization problem. Given P1, we define
	\begin{equation}
	\begin{aligned}
	\label{f_define}
	 f(V_r,E,\Delta\tau)= f(V_r,V_m,E,\Delta\tau) = (V_r+E) \\
	 - (V_d \circ \tau + \sum J_i\Delta\tau_i \epsilon_i) =(V_r+E) - V_m . 
	 \end{aligned}
	\end{equation}
	Then we have:
	%
	\begin{equation}
		\begin{aligned}
			\label{rasl}
			L_\mu(V_r,E,\Delta\tau,Y)= ||V_r||_*+  \lambda*||E||_1 \\ - <Y, f(V_r,E,\Delta\tau)> + \frac{\mu}{2}||f(V_r,E,\Delta\tau)||^2,
		\end{aligned}
	\end{equation}
	where $Y\in\Re^{M\times N}$ is a Lagrange multiplier matrix, $\mu$ is a positive scalar and $<.,.>$ denotes the matrix inner product. For an appropriate choice of the Lagrange multiplier matrix $Y$ and sufficiently large constant $\mu$, it can be shown that  ALM has the same minimizer as that of the original constrained optimization problem.

	\subsection{MeADMM}
	MeADMM is proposed to solve our new problem (P2) with the data lying over a manifold.
	Specifically, we propose to consider $V_r$ as an unknown variable of the optimization by performing variable cloning i.e. $V_{r,i} \to V'_{r,i} \in \mathcal{M}$ and enforcing manifold constraints over the cloned variables ${V'}_r$. This introduces explicitly the manifold constraints at the expenses of replicating a set of variables. 
	The variable cloning $V'_{r,i} = V_{r,i}$ is  used to add the manifold constraint to replace $ V_{r,i}\in\mathcal{M}$ in a set of equations.
	%
	%
	Now, the problem P2 can be rewritten as:
	\begin{myequation}
		\begin{aligned}
		 \hat{V_r},\hat{E},\hat{\tau}=\arg\min  L_\mu(V_r,E,\Delta\tau,Y), \\
			\textit{subject to}  \hspace{0.05in}  V'_{r,i}= V_{r,i}, \hspace{0.05in} V'_{r,i}\in\mathcal{M}
		\end{aligned}
	\end{myequation}
	To solve this problem (P3), we can derive with ADMM the following objective cost function:
	\begin{equation}
		\begin{aligned}
			\label{rasl_manifold}
			L_{\mu,1}(V_r,E,\Delta\tau,Y)= ||V_r||_*+ \lambda*||E||_1 \\
			- <Y, f(V_r,E,\Delta\tau)>+\frac{\mu}{2}||f(V_r,E,\Delta\tau)||^2  \\
			+ \sum^B_i \frac{\sigma_i}{2} ||V_{r,i} - V'_{r,i}||^2
		\end{aligned}
	\end{equation}
	where $\sigma_i$ is a positive value. However, the above objective is still too complicated to be solved, as $L_{\mu,1}$ is the combination of $L_{\mu}$ and another function related to $V_r$ as shown in Eq. \ref{rasl_manifold}. In this case, the calculation of $V_r$ is more complicated than  RASL, which is based on $L_{\mu}$ only. Assuming a linear constraint on $V_m$ and $V_r$, i.e. $ V_m = V_r + E$,  we neglect the error matrix $E$ and obtain the following objective:
	\begin{equation}
		\begin{aligned}
			\label{rasl_manifold1}
			L_{\mu,2}(V_r,V_m,E,\Delta\tau,Y)= ||V_r||_*+ \lambda*||E||_1 \\ 
			-  <Y, f(V_r,V_m,E,\Delta\tau)> +\frac{\mu}{2}||f(V_r,V_m,E,\Delta\tau)||^2 \\
			+  \sum^B_i \frac{\sigma_i}{2} \dot ||V_{m,i} - V'_{m,i}||^2,
		\end{aligned}
	\end{equation}
	with $V'_{m,i}= V_{m,i}$ and $V'_{m,i}\in\mathcal{M}$ as beforehand. 
	The above expression requires a minimization over $V'_{m,i} \in {\mathcal M}$ with $i = 1, \ldots, F$. In \cite{DelBue:etal:PAMI2012}, the manifold constraints are enforced in an ALM strategy by using a matrix projection which computes the solution over several given manifolds (e.g. Stiefel and unit sphere). Differently, in our method data is embedded into a manifold not known a priori but learned from the very same data. 
	
	Now we formalize the manifold by introducing a neighbor-preserving embedding \cite{Varini:KDD,jchen:smc}, which aims to find an estimation of the manifold. Such a formalization is similar to \cite{jchen:smc} which calculates the weights in the process of dimension reduction by LLE \cite{Sam:Science}.
	In particular, the embedding generated by \cite{jchen:smc} is exactly based on a manifold given by a small set of samples.  
	To do so, we find a projection in a manifold based on the ``true" neighbors of input data measured by the Geodesic distance information, thus avoiding the perturbation of samples that are far from the input data.


	\subsection{Embedding for MeADMM}
	
	Let $\mathcal{M}$ be the sample set representing a manifold and $x$ be the embedding of $\mathcal{M}$ via a mapping function $\Phi(\cdot)$.
	\begin{mydef} \label{def1}
		The mapping function $\Phi: x\rightarrow \mathcal{M}$ in the neighbour-preserving embedding method is conducted based on the Geodesic distance, which is defined as follows:
		\begin{enumerate}
			\item
			First we define $\mathcal{M}_1 = \sum_{j=1}^K \mathcal(1-W_{j}) \mathcal{M}_{j}  $ where $W_{j}$ is the Geodesic distance of the sample  $x$ and the $j^{th}$ sample in a  set $\mathcal{M}$. 

			\item
			We define
			$\mathcal{E}= \mathcal{M}-\mathcal{M}_1$, and have:\\ $\Phi_{\alpha,\epsilon}(x,\mathcal{M})= x_{\alpha,\epsilon'} = \mathcal{M}_{1}  + \epsilon' \cdot  S_\alpha[\mathcal{E}]$;
			$ S_\alpha[x]=sign(x)\cdot max\{|x|-\alpha,0\}$\\
			where  $\alpha$ and $\epsilon'$  are used to represent the shrinkage factor and the scaler for reconstruction error respectively. 

			\item
			For a given point projected onto the manifold, larger weights are reasonably assigned to its nearest points in the recovery process.
		\end{enumerate}
	\end{mydef}
	%

	From Definition \ref{def1}, the input sample can be projected onto a manifold via an embedding function by fully exploiting the neighbor structure information \cite{jchen:smc}. As shown in LLE \cite{ruiping}, a local point on a manifold can be represented by a small and compact set of $K$ nearest neighbors  to approximate ISOMAP. In \cite{Varini:KDD}, it has been shown that the Geodesic distance used in ISOMAP is another effective way to locate the neighbors for a linear embedding. We first follow the idea in \cite{tonylin} to estimate the manifold dimension by PCA. Next, we propose the mapping function $\Phi$ to generate an embedding sample that lies on a given manifold. Our idea is similar to \cite{jchen:smc} but the difference lies in its simplicity and feasibility to solve the problem at hand. 
	Note that MeADMM adds an extra computational cost to our problem because variables are added by the cloning mechanism. 	

	The soft-thresholding function for the scalar values \cite{rasl} is defined as: $$S_\alpha[x]=sign(x).\max(|x|-\alpha,0),$$ where $\alpha \ge 0$. When applied to vectors and matrices, the shrinkage operator acts element-wise. Based on Definition \ref{def1}, MeADMM can be alternatively used to solve our problem and in the following we give details about the optimization procedure.

	\begin{center}
		\begin{algorithm} \caption{Main algorithm to solve LRD based on MeADMM}
		\scriptsize
		\label{alg:recovery_algorithm}
			\begin{algorithmic}[1]
				\STATE INPUT:
				\\1)$V_d \circ \tau = [vec(I_1),...,vec(I_B])$, where $I_i$ with $i=1,...,B$ represent $B$ input images;\\ 2) Initialise with $(V_d^0, E^0, \Delta\tau^0).$
				
				\REPEAT
				\STATE  compute Jacobian matrices w.r.t transformations:\\
				$\:\:\:\: J_i \leftarrow \frac{\partial}{\partial \zeta} \left( \frac{vec(I_i \circ \zeta) }{\left\| vec(I_i \circ \zeta)\right\| } \right)\Bigr|_{\zeta = \tau}$;
				
				\STATE
				warp and normalize the images:\\
				$V_d \circ \tau = [\frac{vec(I_1 \circ \zeta) }{\left\| vec(I_1 \circ \zeta)\right\|},\frac{vec(I_2 \circ \zeta) }{\left\| vec(I_2 \circ \zeta)\right\|},...,\frac{vec(I_B \circ \zeta) }{\left\| vec(I_B \circ \zeta)\right\|})$
				
				\STATE
				solve the manifold constraint on the transformation process on
				$V_d \circ \tau + \sum J_i\Delta\tau_i \epsilon_i$ \\
				the details of $V'_{m}$ and $V_r$ are shown in the Alg. 2 and Eq. \ref{rasl_1_vm}. (inner loop)
				\STATE   update the transformation: $   \tau =  \tau + \Delta \tau $ 
				
				\UNTIL some stopping criterion
				\STATE OUTPUT: the solution $(V_r^*,E^*,\Delta\tau^*)$ in our optimization framework.
			\end{algorithmic}
		\end{algorithm}
	\end{center}

	\subsection{The MeADMM algorithm }	
	The main algorithm to solve LRD based on MeADMM is shown in Algorithm 1. In the outer loop, we solve  $\tau$, while other variables such as $V_r$ and $V_m$ are solved in the inner loop (MeADMM).
	A separable structure  based on $L_{\mu,2}(,)$ can be exploited by ADMM, which is:
	\begin{equation}
		\label{others}
		\begin{aligned}
			(V_{r}^{[k+1]},E^{[k+1]},\Delta\tau^{[k+1]},Y^{[k+1]}) = \\ \underset{V_r,E,\Delta\tau}\arg\min  L_{\mu,2}(V_m^{[k]},E^{[k]},\Delta\tau^{[k]},Y^{[k]}).
		\end{aligned}
	\end{equation}
	Details on the solution of Eq. \ref{others} are shown in Algorithm 2. Different from the original objective function in \cite{rasl}, the matrix  $V_m^{[k]}$ needs to be estimated first in order to perform SVD decomposition. Considering the constraint $V_m^{[k]} = V_m^{'[k]}$, the matrix $V_m^{'[k]}$ can be used to replace the original $V_m^{[k]}$ as shown in Algorithm 2. $V_m^{'[k+1]}$ is actually used to approximate $V_m^{[k+1]}$ that lies on the manifold. Now we have a new objective as:
	\begin{equation}
	\small
		\begin{aligned}
			\label{rasl_1}
			L_{\mu,2}(V_r,V_m,E,\Delta\tau,Y)= ||V_r||_*+ \lambda*||E||_1 \\
			 - <Y, ((V_r+E) - V'_m) > +\frac{\mu}{2}||(V_r+E) - V'_m ||^2  \\
			+ \sum^B_i \frac{\sigma_i}{2} \dot ||V_{m,i} - V'_{m,i}||^2, \quad \textit{and} \quad
			V'_{m,i}\in\mathcal{M} 
		\end{aligned}
    \vspace{-0.2em}
	\end{equation}
	

	\begin{center}
		\begin{algorithm} \caption{Variable solution based on the MeADMM algorithm}
		\scriptsize
		\label{alg:recovery_algorithm}
			\begin{algorithmic}[1]
				\STATE  INPUT:
				$V_m^{'{[k]}}$ calculated in Def. 1.

				\STATE compute $(U,\Sigma,\mathcal{V}) \: = \:  \mathcal{SVD}(V_m^{'[k]} +Y^{[k]}/\mu^{[k]}-E^{[k]})$
				\STATE compute $V_{r}^{[k+1]}=US_{\frac{1}{\mu^{[k]}} } |\Sigma| \mathcal{V}^T$
				\STATE compute
				\footnotesize{
					$$ \hspace{-0.2in} E^{[k+1]} =S_{\frac{1}{\mu^{[k]}} } [V_d \circ \tau^{[k]} + \sum J_i\Delta\tau_i^{[k]} \epsilon_i\epsilon_i^T +Y^{[k]}/\mu^{[k]}-V_r^{[k+1]}] $$
					$$ \hspace{-0.2in} \Delta\tau^{[k+1]}=\sum_i J_i (V_{r}^{[k+1]} + E^{[k+1]}- V_{d}^{}\circ \tau^{[k]} -1/(\mu ^{[k]})Y^{[k]})\epsilon_i\epsilon^T_i $$
					$$ \hspace{-0.6in} Y^{[k+1]}=Y^{[k]} + \mu^{[k]} L_u(V_r^{[k+1]},E^{[k+1]},\Delta\tau^{[k+1]},Y^{[k]})$$
					$$ \hspace{-2.1in} \mu^{[k+1]} = \max({0.9 \mu^{[k]},\tilde{\mu}})$$}
				\STATE compute $V_m^{'[k+1]} $ based on Def. 1.
				
				\STATE OUTPUT: the solution $(V_r^*,E^*,\Delta\tau^*, Y^*)$ to the recovery process in our optimization framework.
			\end{algorithmic}
		\end{algorithm}
		\vspace{-0.1in}
	\end{center}

	From Eq. \ref{rasl_1},  the unknowns $Y^{[k+1]}$, $E^{[k+1]}$ and $\Delta^{[k+1]}$ are not directly related to $\sum^B_i \frac{\sigma_i}{2} \dot ||V_{m,i} - V'_{m,i}||^2 $. So we can solve  (Algorithm 2) in a similar method as that of \cite{rasl}. Now only the matrix $V_m^{'[k+1]}$ remains unsolved. Given $V^{[k+1]}_m = V_r^{[k+1]}+E^{[k+1]}$, based on the derivative of Eq. \ref{rasl_1}, $V_m^{'[k+1]}$ is solved as:
	\begin{equation}
		\begin{aligned}
			\label{rasl_1_vm}
			V_m^{'[k+1]} = \mathcal{T}_m  \cdot  V^{[k+1]}_m. 		
		\end{aligned}
	\end{equation}

	As shown in Eq. \ref{rasl_1_vm}, $\mathcal{T}_m$ \footnote{Without considering the manifold constraint, we have  $\mathcal{T}_m =(Y+ (\sigma^*+\mu) \cdot I )^{-1} \cdot (\sigma^*+2 \mu)$, $\sigma_i$ is the diagonal element of $\sigma^*$, and $I$ is the identity matrix.}  is a unknown projection matrix on $V^{[k+1]}_m$. Based on the manifold embedding, the projection is solved in an efficient way, i.e. $V_{m,i}^{'[k+1]} = \Phi_{\alpha,\epsilon}(V_{m,i}^{[k+1]},V_m^{[k+1]})$. 
	The embedding performs well for a small set of nearest samples, which leads to the robustness against severe illumination and corruption. 

	\section{Experiments} \label{sec:experiments}

	%
	%

	We evaluate MeADMM on four datasets including Extended Yale face database B \cite{GeBeKr01:PAMI}, AR face \cite{Martinez:AR}, USPS digits \cite{Hulljj:TPAMI} and planar surfaces (window images) \cite{rasl}. The images in those databases suffer from  rotation, occlusion and lighting variations.  The Geodesic distance is calculated based on $V_r$ with a number of the neighbors ($K$) set to $7$.
	We also empirically set $\alpha =0.05$ and $\epsilon' = 0.85$ in all our experiments.
	
	\begin{figure*}[h!]
		\centering
		\includegraphics[width=1.0\linewidth]{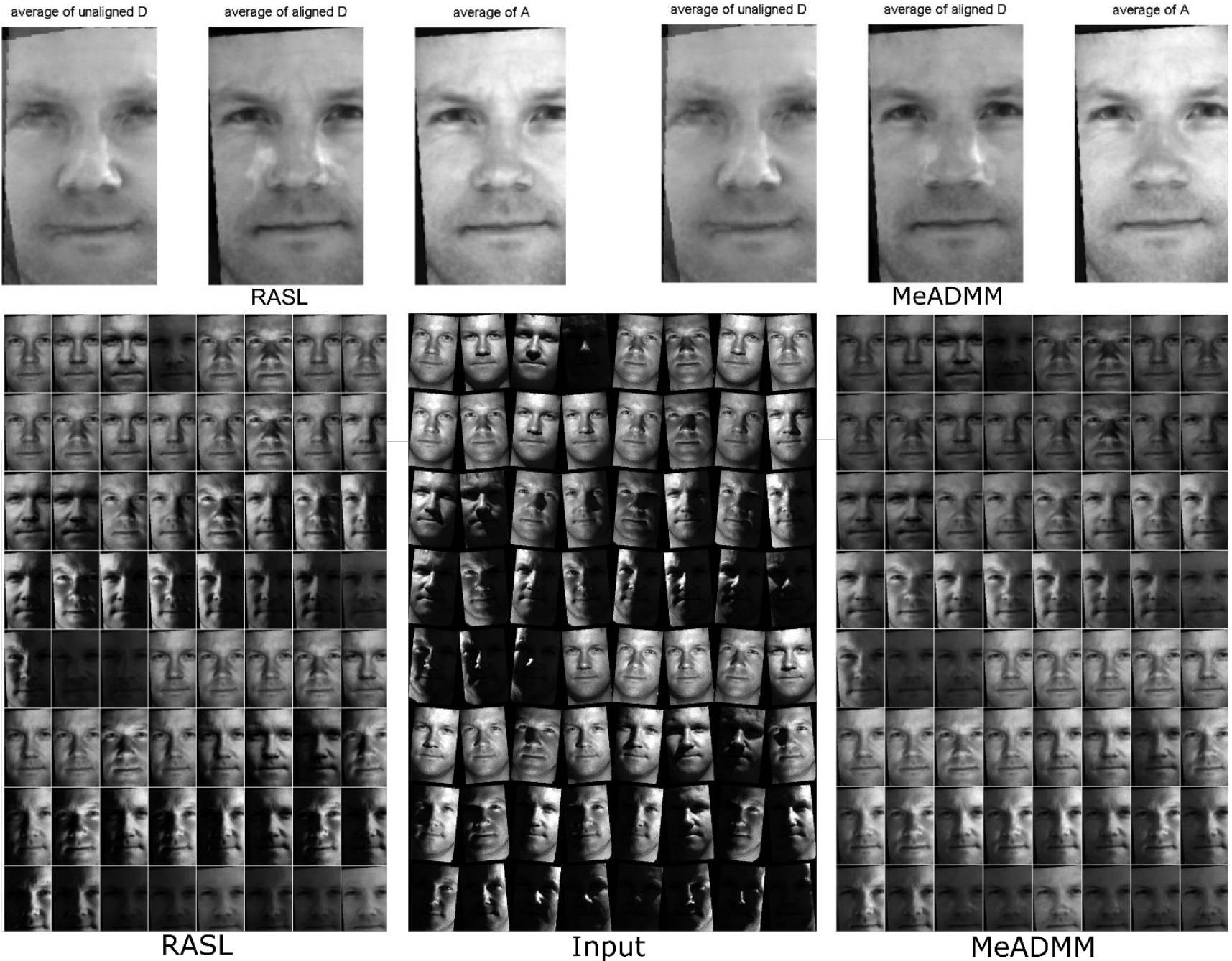}
		
		\caption{Results of RASL and MeADMM on the Extended Yale-B database. In the first row, the averages of input, alignment, and low-rank results are shown for RASL and MeADMM. In the second row, the first and third columns results are obtained by RASL and MeADMM, respectively. (D: Input; A: Low-rank component)}
		\label{fig:yale}
	\end{figure*}
	
	We compare the performance of the proposed MeADMM with RASL \cite{rasl}, which is the state-of-the-art algorithm for LRD. An improved algorithm based on RASL is proposed in \cite{nips13}. Due to lack of implementation, we implemented their algorithm by ourselves, but cannot reproduce the results reported in their paper. Therefore, to facilitate a fair comparison, in this paper we show comparison results of MeADMM and RASL evaluated on each dataset.
	It is also worth noting that the parameter settings of our method are the same as those used by RASL.

	
	\textbf{Extended Yale-B.} In the Extended Yale Face Database B, each subject image contains $64$ different illumination conditions and the images are resized to $42 \times 45$. 
	The 64 images of a subject in a given pose were acquired at 30 frames/second in about 2 seconds, so there is only a small change in head pose and facial expressions in the images. \textbf{To increase the difficulty of the LRD problem, we randomly rotate and shift the face images.} The qualitative results are illustrated in Fig.~\ref{fig:yale}.
	
			\begin{table}[t!]
		\caption{Alignment errors in eye centers, calculated as the distances from
			the estimated eye centers to their ground truth. }
			\small
		\centering
		\begin{tabular}{||c c c c||}
			\hline
			Methods & Mean error (pixel) & Error std. & Max error  \\ [0.5ex]
			\hline
			Initial & 1.69 &  0.428 & 2.23\\
			
			RASL  &   0.16&   0.36  & \bf{1.0}  \\
			
			MeADMM &  \bf{0.14}  & \bf{0.35} & \bf{1.0}  \\
			
			\hline
		\end{tabular}
	\end{table}

	With respect to the alignment, both methods achieve more or less the same performance. For the average faces after alignment, both methods can well solve the misalignment problem. We also present the quantitative results in terms of alignment errors in eye centers on this dataset (see Table 2). As evident in this table, both approaches achieve small misalignment errors for the faces with rotation, lighting variations and shifting.
	Different from RASL that focuses on the misalignment performance, we pay more attention to the recovery effectiveness. Fig.~\ref{fig:yale} shows MeADMM achieving much better performance than RASL in terms of reconstruction quality. Especially for the results on the last three rows,  MeADMM significantly eliminates the illumination variations from the original images, even in the presence of severe illumination changes and misalignment. 

\begin{figure*}[h]
		\centering
		\includegraphics[width=1.0\linewidth]{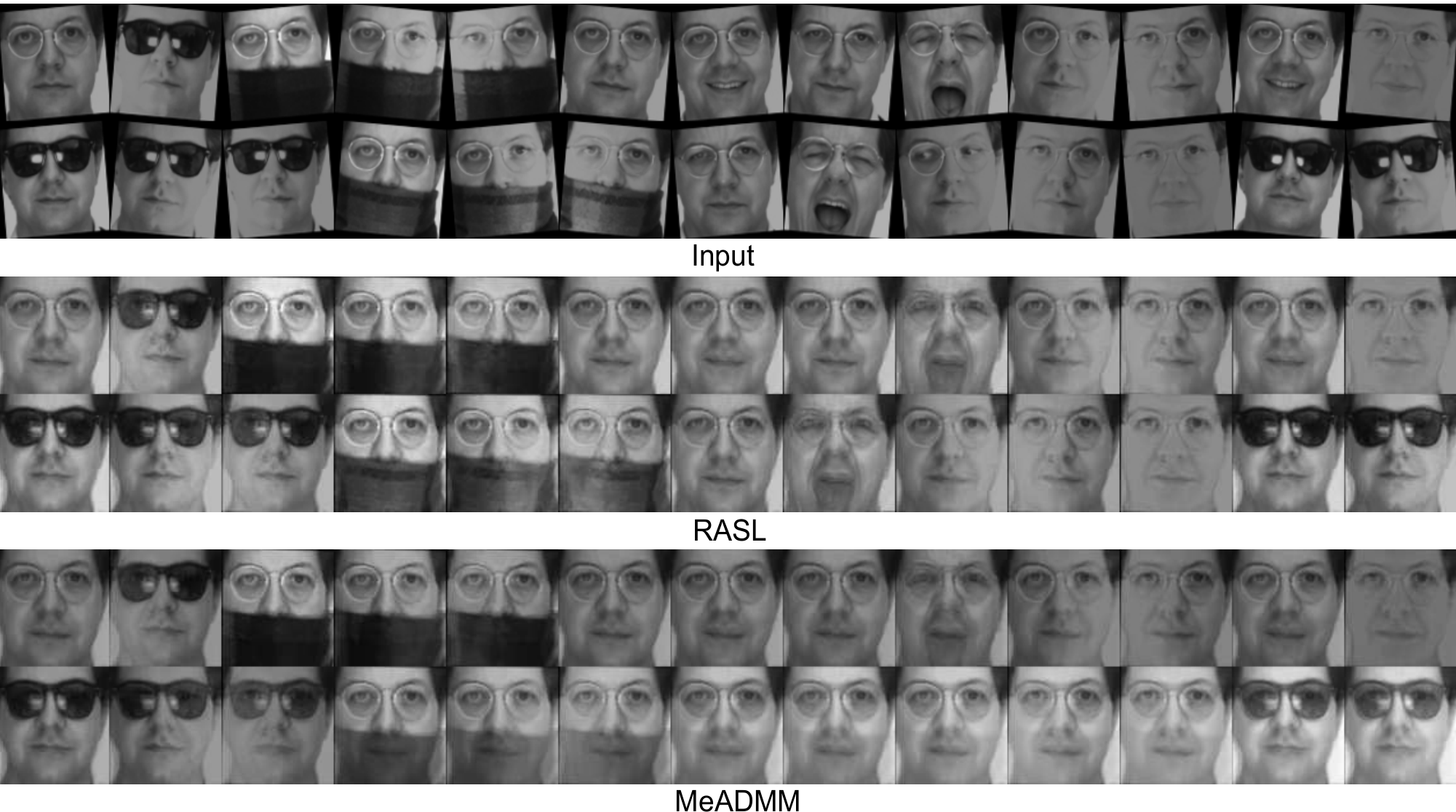}
		
		\caption{Low-rank decomposition results by RASL and MeADMM on the AR database. The top row shows the input images. The second and third rows show the results by RASL and MeADMM, respectively. }
		\label{fig:Ar}
	\end{figure*}

	\textbf{AR Face Database.} We next test MeADMM on the AR face database which contains 126 persons with different facial expressions, illumination conditions, and occlusions (such as sun glasses and scarf). The pictures were taken with no restrictions on participants' appearance (clothes, glasses, etc.), make-up, hair style, etc. in two sessions, separated by two weeks time. For each person, we choose $26$ images ($64 \times 64$) from Session 1 to validate both methods. Different from Yale database B, the faces are severely occluded by glasses and scarf.  It can be observed from  Fig.~\ref{fig:Ar} that MeADMM achieves much  better low-rank images than RASL, especially for the subjects wearing scarf and having large expression variations. The eyes and mouths are almost recovered from the input images as shown in the last row, demonstrating the superiority of MeADMM on image recovery. This is also beneficial for other vision applications such as face recognition and human re-identification. 
	
	\textbf{USPS Database.} In the USPS handwritten digit database, we use a standard subset containing 10-class digit images, and perform MeADMM and RASL on the training set. The low-rank decomposition results are illustrated in Fig. 3. MeADMM successfully recovers all the images of digit 1, but RASL fails on three images. Similarly, MeADMM works well on digit 4, whereas RASL fails to recover one low-rank image. The experiments on the digit images again show the advantages of our method with variations such as rotation, shift and affine transformations. Due to lack of ground truth, we can only show the qualitative results for digits analysis and window image analysis following the evaluation protocol of~\cite{rasl}. 

	\textbf{Planar Surfaces.} Moreover, we demonstrate that MeADMM can be used to align images affected by planar homography transformations. 
	To better demonstrate the robustness of MeADMM, we manipulate the input images by cropping patches or changing illuminations, as in Fig. 4 (input). MeADMM achieves better results than RASL (first and third window images). MeADMM not only aligns the windows and removes the occluded tree branches, but also faithfully inpaints the missing areas. Together with consistent results obtained from faces and digits datasets, we could draw a conclusion that MeADMM is more effective for low-rank calculation, when the data includes rotation, occlusion and illumination variations.
	The performance improvement of MeADMM is attribute to the manifold constraint implicitly learned from the data. 
	
	\textbf{Speed of MeADMM.} Regarding the computational cost, MeADMM is not as fast as RASL due to solving additional variables. Running time for window images is 220ms and 102ms for MeADMM and RASL, respectively on a PC with Intel i5 CPU and 4G RAM. 
	The source code of the proposed MeADMM algorithm will be released to public to facilitate further development.

	\begin{figure*}
    \centering
        \centering
        \includegraphics[width=0.9\linewidth]{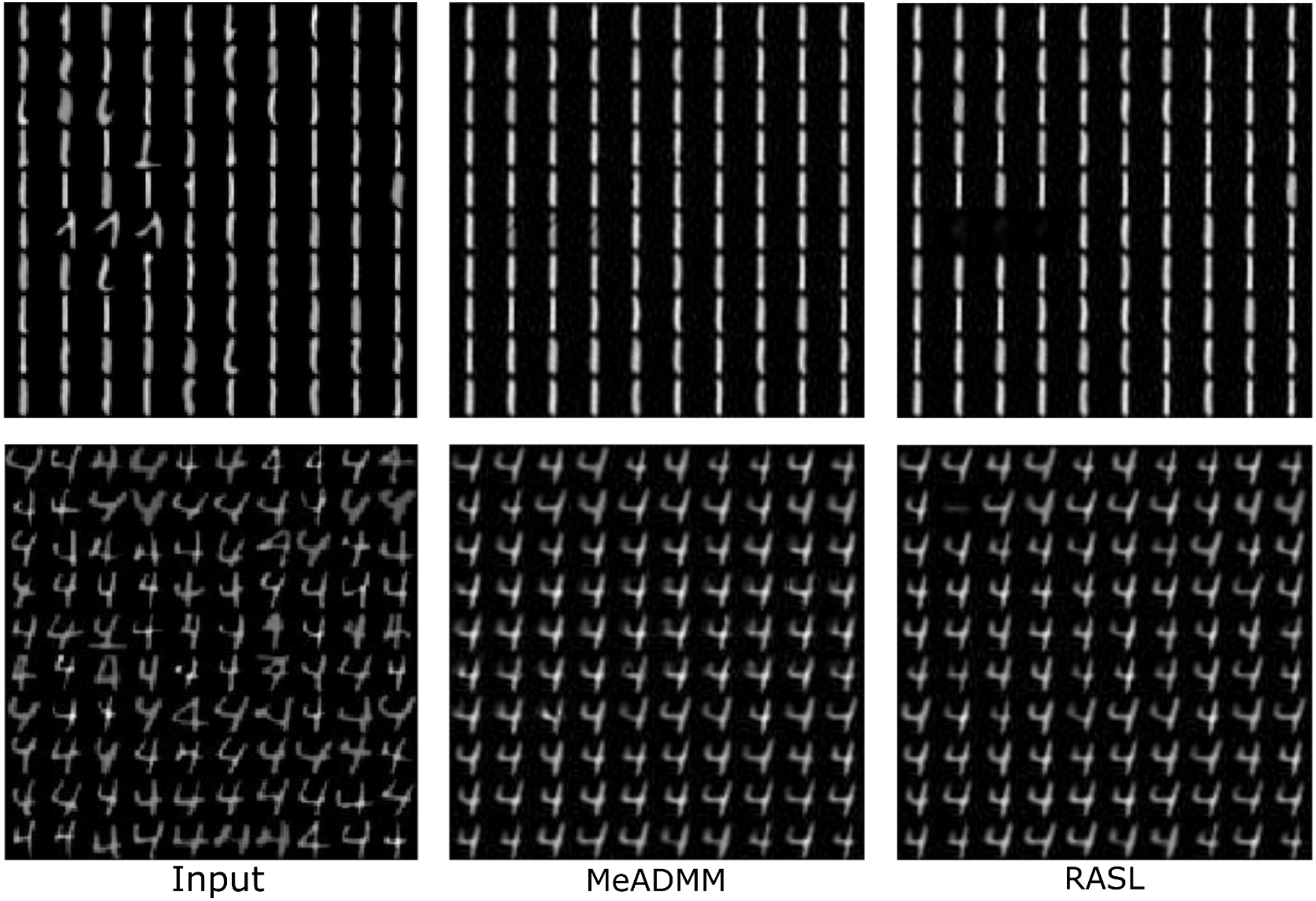} 
        \caption{low-rank decomposition results of MeADMM (second column) and RASL (third column) on the USPS database.}
\end{figure*} 

\begin{figure*}
        \centering
        \includegraphics[width=0.92\linewidth]{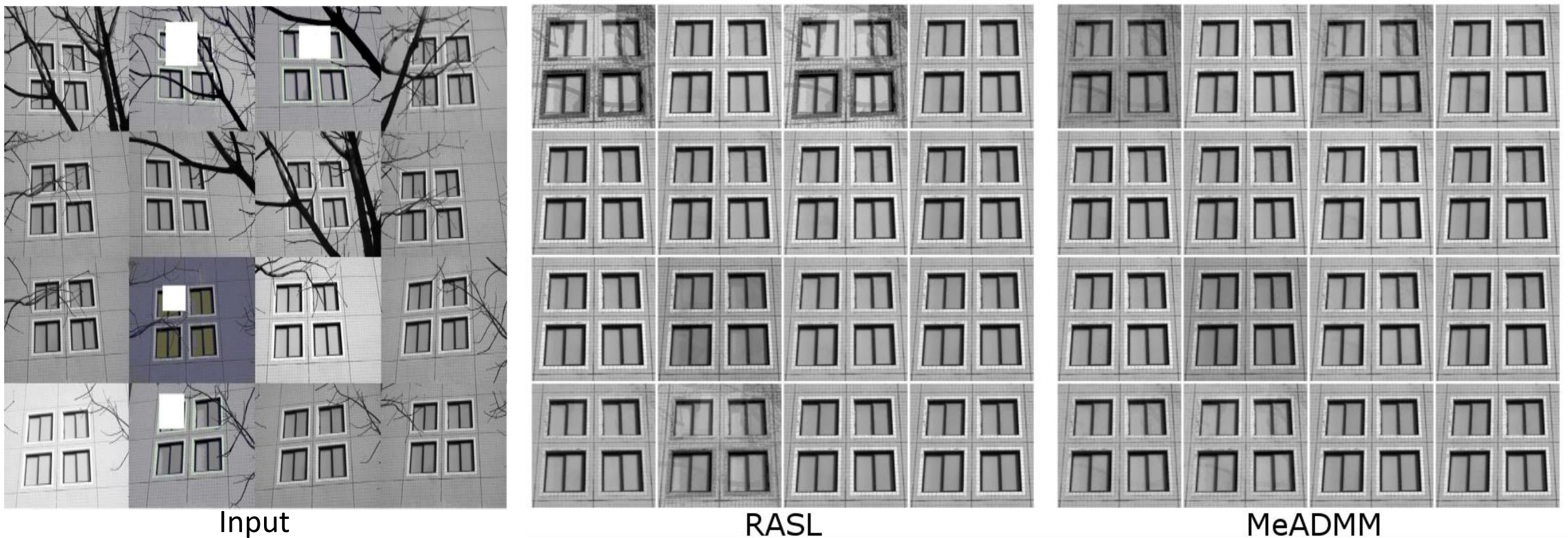} 
        \caption{Low-rank decomposition results of RASL and MeADMM on planar surfaces.}
\end{figure*}

	\section{Conclusion} \label{sec:conclusions}
	
	This paper presents a new insight into low-rank decomposition using manifold constraint and we propose the MeADMM method based on a neighbour-preserving embedding approach. We solve manifold constraint with a projection, which is efficiently calculated during the embedding process. 
	The proposed approach is successfully applied to faces, digits and planar surfaces, showing a consistent increase on image alignment and recovery performance as compared to the state-of-the-art. In future work, we will investigate MeADMM in other applications such as image inpainting, video frame prediction, background substraction, tracking \cite{bc1,bc2,bc3, bc4}.

	\section*{Acknowledgment}	
	The work was supported in part by the Natural Science Foundation of China
	under Contract 61672079 and 61473086.  The work of B. Zhang was supported in part
	by the Program for New Century Excellent Talents University within the
	Ministry of Education, China, and in part by the Beijing Municipal Science
	and Technology Commission under Grant Z161100001616005. Baochang Zhang is the corresponding author.

{\small
\bibliographystyle{ieee}
\bibliography{ijcai17_low_rank}
}

\end{document}